# A Blockchain-Enabled Approach to Cross-Border Compliance and Trust

Vikram Kulothungan
*Capitol Technology University*
North Bergen, U.S.A
vikramk1986@gmail.com

*Abstract*—As artificial intelligence (AI) systems become increasingly integral to critical infrastructure and global operations, the need for a unified, trustworthy governance framework is more urgent that ever. This paper proposes a novel approach to AI governance, utilizing blockchain and distributed ledger technologies (DLT) to establish a decentralized, globally recognized framework that ensures security, privacy, and trustworthiness of AI systems across borders. The paper presents specific implementation scenarios within the financial sector, outlines a phased deployment timeline over the next decade, and addresses potential challenges with solutions grounded in current research. By synthesizing advancements in blockchain, AI ethics, and cybersecurity, this paper offers a comprehensive roadmap for a decentralized AI governance framework capable of adapting to the complex and evolving landscape of global AI regulation.

## I. INTRODUCTION

The rapid advancement and deployment of AI systems across critical sectors have outpaced the development of regulatory frameworks on a global scale. Current regulatory frameworks, while well-intentioned, are fragmented, region-specific, and often unable to keep pace with the rapidly evolving technological landscape. This fragmentation creates significant challenges, particularly for global organizations operating across jurisdictions. Traditional approaches—such as voluntary compliance and geographically constrained regulations—struggle to ensure consistent enforcement, leaving vulnerabilities in security, privacy, and ethical standards.

In response to these challenges, this paper proposes a decentralized AI governance framework that leverages blockchain technology to establish a unified approach. Blockchain's inherent properties, including immutability, transparency, and distributed consensus, uniquely position it to address the complexities of cross-border AI governance. By harnessing blockchain's inherent features, the proposed framework aims to create a secure and ethical environment for AI deployment on a global scale. Traditional governance methods often rely on central authorities or jurisdiction-specific protocols, making them slow to adapt. In contrast, blockchain's decentralized architecture allows for a unified and trustworthy framework that transcends borders, ensuring real-time compliance and continuous auditing of AI systems.

The framework is designed with practical applications in mind, ensuring that it can be implemented effectively across various industries. Additionally, this paper outlines a strategic deployment timeline and addresses key challenges by offering evidence-based solutions, providing a comprehensive roadmap for achieving global AI governance.

## II. BACKGROUND AND RELATED WORK

### A. Current AI Governance Landscape

The global AI governance landscape is a patchwork of varying regulations. The EU's AI Act adopts a risk-based approach with stringent rules for high-risk AI systems, while the U.S. prioritizes innovation through voluntary compliance. This divergence poses challenges for global organizations, leading to gaps in security, privacy, and ethics, and increasing operational risks as companies navigate conflicting regulations.

Taeihagh [1] highlights the need for adaptive regulatory frameworks that evolve with AI advancements across sectors. Veale and Borgesius [2] critique the EU AI Act's implementation, citing the need for consistency across member states and balancing regulation with innovation. These insights underscore the necessity of a unified, flexible AI governance approach, which our framework enables through decentralized technologies for cross-border compliance.

### B. Blockchain in Governance and Compliance

Blockchain, with its immutability, transparency, and decentralized control, offers promise in enhancing governance across sectors, including supply chain management and digital identity. Governatori et al. [3] demonstrate how smart contracts enforce regulatory adherence by embedding legal rules into the blockchain. This concept forms the basis of our proposed compliance system, automating regulatory processes across jurisdictions and reducing the burden on global organizations.

Zetzsche et al. [4] emphasize the need for robust governance in AI-driven financial systems, where blockchain offers transparency and security for monitoring and enforcing compliance. By reducing errors and fraud through a verifiable audit trail, blockchain addresses the unique challenges of high-stakes sectors like finance, which informs our sector-specific implementation scenarios.

### C. AI in Cybersecurity

AI plays a crucial role in cybersecurity, enabling proactive defense through threat detection, anomaly identification, and automated response. However, reliance on AI also introduces vulnerabilities, as AI systems can be targeted or exploited. This dual-edged nature heightens the need for governance frameworks that secure AI systems while preventing misuse.

Brundage et al. [5] examine the malicious uses of AI, outlining scenarios where AI could be weaponized. Their insights inform the security measures in our framework, which ensures resilience against both external attacks and internal misuse, enhancing AI system reliability.

### III. DESIGN PRINCIPLES AND METHODOLOGY

#### A. Design Principles

The design of the framework was guided by several key principles:

- Decentralization: To eliminate single points of failure and ensure global scalability.
- Transparency and Immutability: Blockchain's distributed ledger ensures that all transactions and governance actions are traceable and verifiable.
- Security and Privacy: Ensuring that data exchanges and AI systems remain compliant with data protection laws (e.g., GDPR) and are resilient to cyber-attacks.
- Adaptability: The framework needed to evolve alongside regulatory changes and emerging threats, which is achieved through smart contracts and real-time monitoring.

#### B. Methodology

The framework components were selected to address key governance challenges. A delegated proof-of-stake (DPoS) mechanism facilitates decentralized governance while ensuring scalability and security. Smart contracts automate compliance verification, ensuring real-time regulatory adherence. The decentralized identity system provides tamper-proof identity management, essential for cross-border compliance. These elements form a cohesive framework, integrating governance, compliance, and data security seamlessly.

### IV. PROPOSED DECENTRALIZED AI GOVERNANCE FRAMEWORK

The proposed Decentralized AI Governance Framework "Fig.1", leverages blockchain technology and a risk-based classification system to create a decentralized, transparent, and adaptable governance structure for AI in the industry.

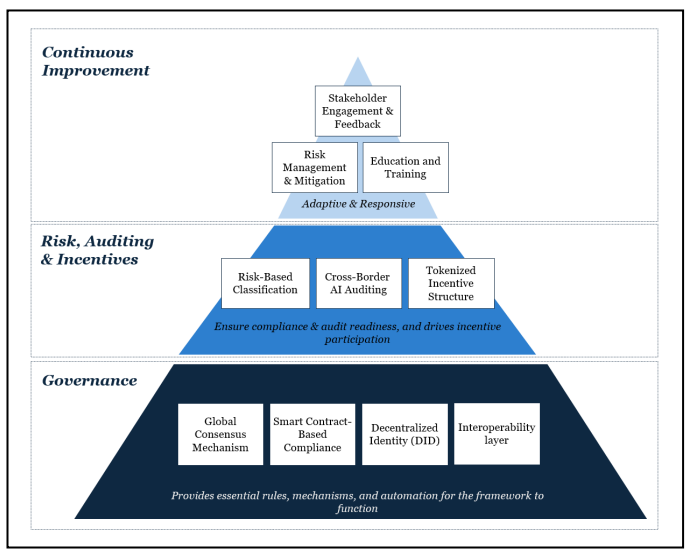

Fig. 1. Proposed Decentralized AI Governance Framework

This framework ensures that AI systems are governed according to their risk levels, addressing the complexities of global financial regulation by integrating key components that provide security, compliance, and trust across international markets.

#### A. Global Consensus Mechanism

This paper proposes a tailored Delegated Proof-of-Stake (DPoS) consensus mechanism designed specifically for the financial sector's unique requirements [6]. This mechanism brings together diverse stakeholders, including major banks, fintech companies, regulatory bodies, and industry experts, to collaboratively establish and update governance standards. These standards are dynamically adjusted based on the classification of AI systems according to their risk levels. This multi-stakeholder approach ensures that governance decisions are informed by a broad spectrum of perspectives, facilitating a more holistic and adaptable governance structure.

*1) Implementation:* The implementation involves active participation from regulators, financial institutions, and AI developers. Given the sector's high stakes, the DPoS mechanism integrates multiple layers of security and accountability:

- Stakeholder Engagement and Voting Power: Stakeholders elect delegates to make governance decisions, with voting power typically proportional to token holdings [7]. To prevent centralization, adjustments such as weighted voting for regulatory bodies and limits on single-entity influence will ensure fair representation for smaller entities.

- Balanced Decision-Making: To ensure a balanced governance structure, the system will implement consensus thresholds and voting caps. These thresholds ensure that decisions require broad support, preventing a small group of delegates from making unilateral decisions. Additionally, an anti-collusion mechanism will be introduced, where the network monitors for patterns of behavior that suggest coordinated voting, triggering additional scrutiny or audits.

- Security Measures: To address DPoS vulnerabilities, the framework integrates multi-signature requirements for critical actions, with approvals needed from multiple delegates. Independent third-party audits will assess governance integrity, reviewing voting power distribution and detecting any anomalies.

- Adaptive Governance Mechanism: The DPoS system will include adaptive governance features that allow the framework to evolve in response to new risks or changes in the regulatory landscape. For example, the system can dynamically adjust the weight of votes or the criteria for delegate selection based on real-time data, such as the emergence of new financial risks or changes in AI-related regulations.

*2) Financial Sector Example:* In practice, this DPoS mechanism enables collaboration between banks, fintechs, and regulators to classify AI systems by risk. For example, AI-driven credit scoring models, deemed high-risk, would be continuously monitored, with governance standards dynamically adjusted based on real-time data. This classification triggers stricter rules, such as increased transparency, mandatory audits, and rigorous compliance checks.

*B. Smart Contract-Based Compliance*

Smart contracts play a pivotal role in automating compliance checks for AI systems, particularly in the highly regulated financial sector. These self-executing agreements ensure adherence to security, privacy, and regulatory standards based on risk classification, adapting as regulations and threats evolve.

*1) Implementation:* Implementing smart contracts involves integrating blockchain and AI to create a dynamic, responsive compliance system. Key components include:

- Modular Smart Contract Design: Smart contracts, developed using languages like Solidity for Ethereum-based systems [8], feature a modular architecture. Each compliance component (e.g., data privacy, risk assessment) can be updated independently, allowing quick adaptation to new regulations without a full contract overhaul [9].

- Integration of AI Capabilities: Natural Language Processing (NLP) algorithms translate legal and regulatory terms into machine-readable code, while machine learning models automate decision-making based on historical compliance data. Continuous learning techniques keep these models updated and effective as new data emerges.

- Predictive Analytics and Real-Time Data Feeds: Smart contracts use predictive analytics to forecast compliance issues and rely on real-time data feeds for current information (e.g., market conditions, regulations). These data feeds are securely integrated into the blockchain, using oracles to bridge the gap between off-chain data sources and on-chain smart contracts.

- Robust Testing and Auditing Framework: Smart contracts undergo formal verification, including model checking and theorem proving, to ensure correct logic. Independent security audits identify and mitigate vulnerabilities like reentrancy attacks and integer overflows.

- Governance and Dispute Resolution Mechanism: A governance mechanism, managed by key stakeholders via a multi-signature wallet, oversees contract updates and dispute resolution. In case of disputes, a decentralized arbitration panel of neutral experts reviews the issue and makes a binding decision.

*2) Financial Sector Example:* In practice, smart contracts can enforce compliance with regulations like Basel III, automatically verifying loan agreements and adjusting terms based on real-time financial data. In derivatives trading, they can execute trades only if they meet compliance criteria, reducing the risk of error, fraud, and non-compliance, while improving efficiency and transparency.

*C. Decentralized Identity for AI systems*

Each AI system in the financial sector will be assigned a unique, tamper-proof decentralized identity (DID) on the blockchain, encapsulating its risk classification and relevant metadata (e.g., system purpose, financial models, compliance status). The DID enhances accountability, allowing comprehensive audits and ensuring AI operates within legal and ethical boundaries.

*1) Implementation:* The implementation of the DID system integrates advanced technologies to ensure security, privacy, and scalability [10]. Key steps include:

- Creation of Unique DIDs: Each AI system is assigned a unique DID using cryptographic algorithms like elliptic curve cryptography (ECC) or RSA, ensuring security and preventing duplication. DIDs are formatted to international standards for interoperability.

- Registration on Permissioned Blockchain: DIDs are registered on a permissioned blockchain that provides privacy and scalability. Only authorized entities access the network, with a Proof of Authority consensus ensuring secure validation of transactions.

- Decentralized Storage Solutions: Metadata (e.g., compliance status, operational history) is stored using decentralized storage solutions like IPFS or Storj, ensuring data availability across multiple nodes. These solutions distribute data across multiple nodes, ensuring that no single point of failure exists and that the data remains accessible even if part of the network goes down. References to this metadata are stored on the blockchain, linking the DID to its associated information.

- Smart Contract Integration: Smart contracts automate DID updates, such as compliance changes or operational modifications. For example, regulatory updates automatically trigger adjustments to the DID, ensuring accuracy and currency.

- Role-Based Access Control (RBAC): RBAC ensures only authorized entities (e.g., regulators, auditors) can view or modify DIDs. The blockchain logs all access, providing an immutable audit trail.

- Application Programming Interfaces (APIs) Integration and Interoperability: APIs enable real-time updates to DIDs, ensuring changes in AI status or compliance are immediately reflected. The system is designed for interoperability with blockchain networks and traditional financial systems, using standards like ISO 20022.

- Audit Trail and Compliance Monitoring: The blockchain records an immutable audit trail of DID changes, enabling auditors to track AI operations and compliance. Oracles provide real-time regulatory data, ensuring the DID reflects the system's current compliance status.

*2) Financial Sector Example:* In the financial sector, high-risk AI models, such as trading algorithms, will receive blockchain-based DIDs that include risk classification and compliance status. If compliance changes due to a regulatory update, smart contracts automatically update the DID and trigger necessary audits, ensuring continuous oversight.

*D. Cross-Border AI Auditing*

A decentralized network of accredited AI auditors will verify compliance across jurisdictions, focusing on high-risk systems defined by respective AI regulations. This approach ensures AI systems comply with international financial regulations through a consistent and transparent cross-border auditing process.

*1) Implementation:* The implementation of a decentralized AI auditing network is a multi-layered process that requires the integration of advanced technologies and rigorous governance protocols. Key components include:

- Rigorous Accreditation Process for Auditors: The success of the auditing network relies on a robust accreditation process. Auditors undergo stringent training to ensure expertise in AI and financial regulations. Accreditation bodies, recognized nationally and internationally, certify auditors using smart contracts, which automatically verify qualifications and issue blockchain-recorded certifications [11].

- Permissioned Blockchain Infrastructure: Audit results are recorded on a permissioned blockchain, accessible only by authorized parties (e.g., regulators, auditors), ensuring confidentiality. The immutable ledger provides a tamper-proof record for future reference and regulatory review.

- Smart Contract Automation: Smart contracts automate audits, including scheduling, evidence submission, and certification of compliance. High-risk AI systems undergo more frequent audits, triggered by their risk classification, ensuring consistent procedures across jurisdictions.

- Standardized Audit Protocols and Consistency: To address the challenges of cross-jurisdictional audits, standardized audit protocols will be developed and codified within the smart contracts. These protocols will align with international best practices and be adaptable to the specific legal and regulatory requirements of different regions. A decentralized body governs regular updates to ensure alignment with evolving laws and technologies [12].

- Security and Data Privacy Measures: Zero-knowledge proofs (ZKPs) allow auditors to verify compliance without accessing sensitive data, ensuring privacy. The blockchain uses encryption and multi-signature access controls to prevent unauthorized access to audit records.

- AI-Powered Continuous Monitoring: AI-powered monitoring tools will be integrated into the auditing network to provide continuous compliance assessments. These tools can analyze large volumes of data in real-time, identifying potential compliance issues before they escalate into significant risks. By leveraging machine learning algorithms, the system can also predict future compliance challenges based on historical data and emerging trends, allowing for proactive governance and timely audits.

*2) Financial Sector Example:* In the financial sector, AI models in high-risk applications, such as credit scoring and algorithmic trading, undergo regular audits by accredited auditors. For example, a global AI-driven credit scoring system is periodically audited for compliance with local and international regulations. Audit findings can be recorded on the blockchain, providing an immutable, transparent record accessible by regulators across jurisdictions, ensuring continuous compliance.

*E. Tokenized Incentive Structure*

A native token will be introduced to incentivize participation in compliance processes and reward AI systems that maintain high standards in the financial sector. This tokenized system fosters a self-sustaining ecosystem where compliance is both mandated and economically rewarded [13]. Stakeholders, such as financial institutions and auditors, can stake tokens to influence governance and earn rewards based on the compliance performance of AI systems, with higher rewards for systems classified as high-risk.

*1) Implementation:* The implementation of the tokenized incentive structure includes the following key components:

- Creation of the Native Token: The native token, developed using blockchain technology, will be secure, transparent, and transferable. A finite supply will prevent inflation, with allocations for rewards, governance, and development. The token's value will be tied to its utility in the compliance ecosystem (e.g., reduced fees, governance influence).

- Compliance Metrics and Smart Contracts: Smart contracts will automate token distribution based on predefined compliance metrics derived from audit results and AI-powered monitoring. For example, compliance with Basel III could be measured by the accuracy of AI risk assessments, with tokens awarded for strong performance. Smart contracts will ensure that rewards are distributed transparently and fairly, reducing the potential for human error or bias.

- Staking Mechanism and Governance: Stakeholders can stake tokens to influence governance and earn rewards based on staking duration and contributions. Safeguards like voting caps or quadratic voting will prevent the concentration of power.

- AI-Powered Monitoring and Continuous Assessment: AI algorithms will continuously monitor AI systems' compliance, directly influencing token distribution. Additionally, AI-powered tools will be used to identify potential gaming of the system or other forms of manipulation, triggering audits or penalties as necessary.

- Legal and Regulatory Considerations: Legal and regulatory considerations, such as security classification, tax implications, and AML compliance, will be addressed. Experts will ensure the token meets all relevant laws, with mechanisms for transparency and traceability of transactions.

*2) Financial Sector Example:* In the financial sector, high-risk AI systems, such as those in credit scoring or algorithmic trading, that comply with regulatory standards will earn tokens as rewards. For instance, a Basel III-compliant credit scoring system may receive tokens, offering benefits like reduced fees or access to premium services. These tokens can also be staked to influence future governance decisions.

### F. Interoperability Layer

The interoperability layer ensures seamless integration of high-risk AI systems with financial platforms and blockchain networks, enabling secure data exchanges and collaboration. It supports efficient operation by governing interactions with stringent security and compliance standards.

*1) Implementation:* The implementation of the interoperability layer involves the development of a comprehensive architecture that facilitates the integration of high-risk AI systems with existing financial systems and multiple blockchain platforms. Key components include:

- Development of Standardized Protocols: Standardized protocols will ensure secure and compliant data exchanges between AI systems and financial platforms, handling transaction data, risk assessments, and compliance reports. Based on standards like ISO 20022, these protocols will adapt to regulatory changes.

- Creation of Robust APIs and SDKs: APIs and SDKs will enable secure integration of AI systems with the interoperability layer [14], supporting functions like data retrieval, compliance verification, and audit logging. SDKs will ensure compatibility across systems, helping developers easily integrate their AI systems.

- Implementation of Blockchain Connectors and Data Transformation Services: Blockchain connectors will ensure cross-chain compatibility, allowing AI systems to interact with multiple blockchain networks. Data transformation services will convert information between formats, facilitating integration between legacy systems and modern platforms.

- Encryption and Authentication Mechanisms: Data exchanges will be secured using encryption and authentication mechanisms like public-key cryptography, multi-factor authentication, and TLS (Transport Layer Security) for data in transit, with advanced encryption standards (AES) protecting data at rest.

- Automated Compliance Checking Tools: Automated compliance tools will monitor data exchanges for regulatory adherence, flagging issues in real-time and enabling immediate corrective actions.

- Scalability and Layer-2 Scaling Solutions: The architecture of the interoperability layer will be designed for scalability, capable of handling high volumes of transactions and data exchanges [15]. The interoperability layer will use Layer-2 scaling solutions (e.g., state channels, sidechains) to manage high-frequency compliance checks off-chain, reducing the load on the main blockchain and maintaining security during large-scale operations.

- Comprehensive Monitoring and Logging Systems: Real-time monitoring and logging systems will track all interactions, compliance checks, and transactions. Logs will be securely stored on the blockchain, providing an immutable record for audits and regulatory reviews.

- Version Control and Backward Compatibility: Version control will manage updates, track changes, and maintain backward compatibility, ensuring AI systems and financial platforms operate smoothly. It will also facilitate collaboration among developers for ongoing improvements.

*2) Financial Sector Example:* In the financial sector, the interoperability layer enables high-risk AI systems, such as trading algorithms or credit scoring systems, to securely interact with exchanges, banks, and regulatory platforms. Standardized APIs and blockchain connectors ensure compliance with data protection and facilitate cross-border transactions.

### G. Stakeholder Engagement and Feedback

A continuous stakeholder engagement and feedback mechanism will be established to ensure the governance framework remains adaptive and effective, focusing on high-risk AI systems and updating standards to meet evolving financial sector needs [16].

*1) Implementation:* The stakeholder engagement mechanism will ensure all relevant voices are included in the governance process. Key components include:

- Stakeholder Roles and Responsibilities: The framework will define the roles of regulators, banks, fintechs, AI developers, and industry experts. Regulators focus on legal alignment, while banks and fintechs provide industry insights, and AI developers offer technical expertise. Each group will have specific channels for providing input.

- Diverse Engagement Methods: Engagement methods will include regular forums, bi-annual workshops, and an online platform for real-time input. The platform will feature feedback forms, discussion forums, and voting to prioritize issues based on stakeholder consensus.

- Rotating Stakeholder Advisory Board: A rotating stakeholder advisory board, with representatives from each group, will provide ongoing guidance and recommend governance updates. The board's recommendations will be documented and made publicly available to ensure transparency.

- Annual Governance Review and Transparency Reports: An annual governance review will incorporate stakeholder feedback and industry trends into updated standards. Transparency reports will detail how feedback has influenced updates, including examples of changes made.

- Collaborative Working Groups: Collaborative working groups will be formed to address complex challenges, such as fairness in AI decisions. The groups will operate on a project basis, and their findings will be integrated into the governance framework.

- Regulatory Liaison Program: A regulatory liaison program will facilitate communication between governance administrators and regulators, ensuring alignment with evolving legal requirements. Regulators will also provide direct feedback and suggest improvements.

- Metrics for Measuring Engagement Effectiveness: Metrics will measure engagement effectiveness, including participation levels and actionable feedback submissions. Regular assessments will identify gaps and ensure the framework evolves as needed.

*2) Financial Sector Example:* In the financial sector, quarterly forums will allow stakeholders to discuss AI governance improvements, focusing on high-risk systems like trading models or credit scoring. Insights from these discussions will inform updates to governance standards, ensuring fairness and transparency.

### H. Risk Management and Mitigation

Risk management is integral to the governance framework, ensuring high-risk AI systems remain secure, reliable, and compliant. The framework integrates continuous assessments, proactive mitigation, and dynamic updates to governance standards [17].

*1) Implementation:* Risk management strategies in the governance framework leverage advanced technologies and cross-functional collaboration [18]. Key components include:

- Continuous, Automated Risk Assessments: The framework will use continuous, automated risk assessments through smart contracts to identify vulnerabilities and emerging threats. Smart contracts will trigger compliance checks and mitigation actions in real-time if abnormal behavior or risks are detected.

- Regular Vulnerability Testing: High-risk AI systems will undergo regular vulnerability testing, covering both software and hardware. Results will inform compliance checks and governance updates, with additional tests triggered by significant system changes.

- AI-Powered Monitoring Tools: AI-powered tools will continuously monitor AI systems using machine learning to detect risks in real-time. Predictive analytics will anticipate future risks based on historical data, allowing preemptive action before risks materialize.

- Automated Compliance Checks and Dynamic Governance Updates: The framework will dynamically update standards in response to new risks. Automated compliance checks, triggered by risk assessments, will ensure high-risk AI systems remain compliant with evolving regulatory requirements.

- Incident Response Protocols and Stakeholder Communication: Incident response protocols will manage risks, outlining steps for security breaches or system failures. Incident response teams will quickly resolve issues, with a communication system to keep stakeholders informed of actions taken.

- Cross-Functional Collaboration: Risk management will involve collaboration between risk teams, compliance officers, and AI developers. Regular cross-functional meetings will review risk assessments and coordinate responses to emerging threats.

- Metrics for Assessing Risk Management Effectiveness: Metrics will assess risk management effectiveness, including identified vulnerabilities, incident resolution times, and compliance rates. Regular assessments and stakeholder feedback will help improve and adapt the framework.

*2) Financial Sector Example:* In the financial sector, high-risk AI systems like trading models or credit scoring algorithms will undergo regular vulnerability testing. Test results will trigger compliance checks, security measures, and updates to governance standards, ensuring AI systems remain secure and compliant.

### I. Education and Training

Comprehensive education and training resources will be provided to ensure stakeholders understand the governance framework and can manage high-risk AI systems responsibly within the financial sector.

*1) Implementation:* The implementation of education and training resources involves several key components:

- Structured Curriculum Development: A structured curriculum will cover AI fundamentals, governance frameworks, and legal and ethical considerations. It will include role-specific modules for executives, technical staff, and compliance officers, ensuring tailored, relevant training.

- Multi-Format Learning Materials: The program will offer multi-format learning materials, including interactive online modules, video-based training, hands-on workshops, and simulations of high-risk AI scenarios to provide practical experience.

- Certification Programs: Certification programs, including the AI Governance Professional (AIGP) certification, will align with EU AI Act requirements and serve as a mark of expertise. Certification will require completion of role-specific modules and assessments, with periodic recertification to maintain up-to-date knowledge.

- Continuous Learning Platform: A continuous learning platform will offer updated materials, including regulations, case studies, and new training modules. Regular expert-led workshops and webinars will dive into emerging topics and complex challenges.

- Practical Application through Case Studies and Simulations: Case studies and simulations based on real-world financial scenarios, such as AI-driven credit scoring or algorithmic trading, will provide hands-on learning. Simulations will allow participants to apply knowledge in a risk-free environment and discuss outcomes in group settings.

- Collaborative Learning Forums: Collaborative learning forums will be hosted on the platform for peer-to-peer knowledge sharing, moderated by experts to encourage active participation and discussion on AI governance.

- Regular Assessments and Feedback Mechanisms: Regular assessments will measure participants' understanding, and feedback collected via surveys will help refine the curriculum and update learning materials as needed.

*2) Financial Sector Example:* In the financial sector, workshops and training sessions will help banks and fintechs integrate AI governance practices, particularly for high-risk systems like automated lending or fraud detection. Certification programs will ensure AI risk managers are equipped with the skills needed for responsible AI management.

## V. EUROPEAN UNION ARTIFICIAL INTELLIGENCE ACT (EU AI ACT) ALIGNMENT

The EU AI Act is recognized as one of the most comprehensive and influential regulatory frameworks for AI governance worldwide. Its focus on high-risk AI systems, transparency, and accountability sets a global standard for ethical AI deployment. Ensuring alignment with this Act is critical for building trust in AI technologies, particularly in sectors like finance where the risks and stakes are high.

The "TABLE I." below highlights how the key components of the proposed decentralized AI governance framework align with the EU AI Act's stringent requirements. By demonstrating regulatory compliance through mechanisms such as risk classification, automated compliance, decentralized identities, and cross-border auditing, the framework provides a robust, adaptable solution for AI governance that can be applied across international markets.

TABLE I. EU AI ACT ALIGNMENT

| Component | EU AI Act Requirement | Framework Alignment |
|---|---|---|
| Risk Classification (Global Consensus Mechanism) | Classify AI systems based on risk levels (e.g., high-risk systems) | The framework uses a Delegated Proof-of-Stake (DPoS) consensus mechanism to classify AI systems by risk. High-risk AI systems are subject to stricter compliance and governance. |
| Compliance Automation (Smart Contracts) | High-risk AI systems must undergo regular conformity assessments | Smart contracts automatically enforce compliance by conducting real-time assessments of AI systems, ensuring adherence to regulatory standards without manual oversight. |
| Decentralized Identity (DID) | Unique identifier for traceability of AI systems | Each AI system is assigned a decentralized identity (DID), providing traceability throughout its lifecycle. This ensures transparent audit trails for regulators. |
| Cross-Border Auditing | Regular audits by accredited third-party bodies | The decentralized auditing network ensures that accredited auditors perform regular compliance checks on high-risk AI systems across different jurisdictions. |
| Ethics and Fairness | Maintain fairness, accountability, and transparency | Ethical concerns such as bias and privacy are mitigated through continuous monitoring and decentralized identities, ensuring the framework remains aligned with EU AI ethical standards. |
| Interoperability | Ensure interoperability across systems and regulatory environments | The interoperability layer supports seamless integration with various financial platforms, ensuring compliance with cross-border data protection and security requirements. |

## VI. ETHICAL CONSIDERATIONS AND POTENTIAL CONFLICTS

The governance framework includes comprehensive education and training resources focused on managing high-risk AI systems as defined by the AI Act. These resources ensure that stakeholders understand the framework, its principles, and applications, enabling responsible AI management within the financial sector.

*1) Bias in AI Decision-Making:* AI systems in finance are prone to biases from training data, potentially leading to discriminatory decisions like unfair credit scoring or biased investment strategies. The framework's decentralized governance mechanisms will monitor and audit for bias using ethical auditing through smart contracts, ensuring regular fairness evaluations. Aligning global ethical standards remains a challenge, as bias prevention must account for varying cultural perspectives.

*2) Data Privacy vs Transparency:* Blockchain's transparency can conflict with data privacy regulations like GDPR, particularly when high-risk AI systems process sensitive data. To resolve this, the framework stores only essential metadata on-chain and uses decentralized storage (e.g., IPFS) for personal data. This ensures compliance with privacy regulations while maintaining necessary transparency for governance.

*3) Accountatbility in a Decentralized System:* Decentralized governance distributes decision-making among stakeholders, enhancing fairness but raising accountability concerns when AI systems make harmful decisions. The framework uses Decentralized Identities (DID) for AI systems and stakeholders, ensuring traceability of actions and enforceable accountability.

*4) Tokenized Incentives and Governance Manipulation:* While tokenized incentives encourage participation and reward ethical compliance, they risk creating profit-driven governance, where powerful entities might manipulate decisions. The framework mitigates this with anti-collusion measures, such as weighted voting and caps on voting power, to prevent undue influence. Real-time audits will also detect unusual voting patterns to safeguard against manipulation.

*5) Cross-Cultural Ethical Disparities:* AI ethics vary across regions, creating challenges in forming universal standards. For instance, data privacy in Europe is stricter than in other regions. The decentralized framework will aim to balance consistent ethical standards with flexibility to accommodate cultural differences, while maintaining core principles of fairness, transparency, and accountability.

## VII. Implementation Timeline and Geographical Considerations

We propose a phased implementation of the AI governance framework over the next decade, accounting for diverse regulatory environments, technological infrastructures, and cultural contexts to ensure adaptability and scalability.

*1) Phase 1 (2024-2026): Development and Testing of Blockchain Infrastructure and Smart Contracts:* The primary goal of Phase 1 is to build and test the foundational elements, including blockchain infrastructure and smart contracts..

- Blockchain Infrastructure Development: Ensure security, scalability, and compliance with financial sector requirements.
- Smart Contract Design and Testing: Test governance functions like compliance verification and risk management.
- Regulatory Alignment: Collaborate with EU and US regulators to align with legal standards.
- Pilot Testing: Initial tests in select EU and US financial institutions, focusing on high-risk AI systems.
- Geographical Focus: EU and US markets provide a robust environment for initial testing.

*2) Phase 2 (2026-2028): Pilot Programs and Expansion to Key Markets:* Phase 2 aims to expand the framework's reach by launching pilot programs in additional sectors and markets, testing the framework's interoperability, compliance, and risk management features across diverse regulatory environments.

- Sector Pilots: Test in the financial and healthcare sectors, focusing on high-risk AI (e.g., lending algorithms, AI diagnostics).
- Interoperability Testing: Ensure seamless operation across diverse regulatory and technological environments.
- Regulatory Engagement: Engage with regulators in Canada, Japan, and Australia.
- Stakeholder Training: Roll out comprehensive training programs in these regions.
- Geographical Focus: Expand to Canada, Japan, and Australia in addition to the EU and US.

*3) Phase 3 (2028-2030): Full Implementation Across G20 Countries:* The goal of Phase 3 is to achieve full implementation of the governance framework across G20 countries, with a particular emphasis on integrating emerging economies.

- Adaptation to Local Contexts: The framework will be adapted to fit the specific regulatory, technological, and cultural contexts of each G20 country, ensuring that it remains effective and relevant across different environments.
- Partnership Development: Form strategic partnerships with local institutions and regulators.
- Continuous Monitoring and Feedback: Implement systems to track progress and integrate stakeholder feedback.
- Capacity Building: Ongoing capacity-building initiatives will be introduced to support the adoption and operation of the framework in emerging economies.
- Geographical Focus: Full implementation will be pursued across all G20 countries, with specific efforts to integrate emerging economies into the global governance framework.

*4) Phase 4 (2030 Onwards): Global Rollout and Continuous Refinement:* Phase 4 will focus on extending the framework's reach beyond the G20 to all major economies, with a continuous refinement process to adapt to new technological developments, regulatory changes, and stakeholder feedback.

- Global Rollout: Extend the framework to all major economies, ensuring adaptability to local conditions.
- Continuous Refinement: Update the framework based on technological advancements, regulatory changes, and stakeholder feedback.
- Global Collaboration and Knowledge Sharing: Establish an international consortium for collaboration and knowledge sharing.
- Final Integration: The final phase will focus on ensuring that the framework is fully integrated into global financial and technological ecosystems, with standardized protocols and practices that enable seamless operation across borders.
- Geographical Focus: This phase will expand the framework's reach beyond the G20, targeting all major economies and ensuring that the framework is adaptable to various global contexts.

## VIII. Conclusion

The proposed decentralized AI governance framework marks a major advancement in addressing the security, privacy, and trust challenges of AI systems on a global scale. By leveraging blockchain technology, it offers an innovative solution to the current limitations of AI governance, particularly in the financial sector, where high-risk AI applications are increasingly prevalent.

The framework's key components—including the global consensus mechanism, smart contract-based compliance, decentralized identity for AI systems, and tokenized incentives—work together to create a transparent, adaptable, and inclusive system. This approach fosters innovation while maintaining rigorous safety and ethical standards, aligning with regulations such as the EU AI Act. The focus on stakeholder engagement, risk management, and education further strengthens the framework's long-term potential for success and widespread adoption.

The phased implementation plan (2024–2030 and beyond) provides a practical roadmap for gradual adoption and refinement. By accommodating geographical differences and integrating emerging economies, the plan emphasizes inclusivity and global reach, ensuring the framework remains effective as it expands worldwide.

As AI becomes more integral to society and the economy, this comprehensive governance framework will be essential in building trust among stakeholders and unlocking AI's full potential. Its proactive approach to managing risks and fostering ethical AI development is particularly critical in high-stakes sectors like finance, where governance failures can have severe consequences.

While challenges in global coordination and regulatory alignment persist, this framework provides a solid foundation for addressing these issues. Ongoing research and development in AI governance will be crucial to refining the framework, addressing gaps, and enhancing adaptability to emerging technologies and evolving regulations.

In conclusion, the framework sets a high standard for responsible AI development, paving the way for AI's safe and ethical use across the global economy.